\documentclass[10pt, a4paper, onecolumn]{article}

\usepackage[numbers]{natbib}

\usepackage[T1]{fontenc}
\usepackage{lmodern}

\usepackage{natbib}
\usepackage[utf8]{inputenc} 
\usepackage{booktabs}       
\usepackage{amsfonts}       
\usepackage{nicefrac}       
\usepackage{microtype}      
\usepackage{xcolor}         
\usepackage{graphicx}
\usepackage{amsmath,amssymb,amsfonts}
\usepackage{bm}

\newcommand{\R}{\mathbb{R}}

\title{Maximum and Leaky Maximum Propagation}

\author{
	Wolfgang Fuhl
	Department of Human Computer Interaction\\
	University Tübingen\\
	Tübingen, 72076 \\
	\texttt{wolfgang.fuhl@uni-tuebingen.de} \\
}

\begin{document}
	
	\maketitle
	
	\begin{abstract}
		In this work, we present an alternative to conventional residual connections, which is inspired by maxout nets. This means that instead of the addition in residual connections, our approach only propagates the maximum value or, in the leaky formulation, propagates a percentage of both. In our evaluation, we show on different public data sets that the presented approaches are comparable to the residual connections and have other interesting properties, such as better generalization with a constant batch normalization, faster learning, and also the possibility to generalize without additional activation functions. In addition, the proposed approaches work very well if ensembles together with residual networks are formed. Code link:  https://atreus.informatik.uni-tuebingen.de/seafile/d/8e2ab8c3fdd444e1a135/?p=\%2FMaximumPropagation\&mode=list
	\end{abstract}

	\section{Introduction}
	The development of deep neural networks~\cite{rosenblatt1958perceptron,lecun1989backpropagation,he2016deep,AAAIFuhlW,NNPOOL2020FUHL,NORM2020FUHL,RINGRAD2020FUHL,RINGRAD2020FUHLARXIV} has undergone steady advancements over the past few decades. These advancements, especially through the introduction of convolutions~\cite{lecun1989backpropagation}, residual blocks~\cite{he2016deep}, and memory functionality~\cite{hochreiter1997long}, have led to deep neural networks being the de facto standard approach for many areas of algorithm development today. This has led to major advances in the areas of security~\cite{RLDIFFPRIV2020FUHL,GMS2021FUHL}, computer vision~\cite{hassaballah2020deep,WTCKWE092015,WTTE032016,062016,CORR2017FuhlW2,CORR2016FuhlW,WDTE092016,WTCDAHKSE122016,WTCDOWE052017,WDTTWE062018,VECETRA2020,ETRA2018FuhlW,ETRA2021PUPILNN,WTDTWE092016,WTDTE022017,WTE032017,WF042019,ICCVW2019FuhlW,CAIP2019FuhlW,ICCVW2018FuhlW}, speech recognition~\cite{lu2020automatic}, pattern recognition~\cite{bayrak2020deep,UMUAI2020FUHL,C2019,FFAO2019}, validation~\cite{ICMV2019FuhlW,NNVALID2020FUHL}, human computer interaction~\cite{brodic2020human,NNETRA2020,CORR2017FuhlW1,FCDGR2020FUHLARX,FCDGR2020FUHL,fuhl2018simarxiv,ICMIW2019FuhlW1,ICMIW2019FuhlW2,EPIC2018FuhlW,MEMD2021FUHL,MEMD2021FUHLARX}, perception understanding~\cite{AGAS2018,ROIGA2018,ASAOIB2015}, and big data processing~\cite{amanullah2020deep,ICML2021DS}. The application areas of deep neural networks in modern times are Autonomous Driving~\cite{janai2020computer}, Gaze Estimation~\cite{yu2020unsupervised}, Collision Detection~\cite{fan2021learning}, Industrial Algorithm Development~\cite{liang2020toward}, Tumor Detection~\cite{saba2020brain}, Person Identification~\cite{ye2021deep}, Text Translation~\cite{popel2020transforming}, Image Generation~\cite{goodfellow2014generative}, Quality Enhancement of Images, and many more. 
	
	The research areas of deep neural networks include activation functions~\cite{hayou2019impact}, optimization methods~\cite{kingma2014adam}, robustness of neural networks~\cite{madry2017towards}, explainability and definitional domains~\cite{samek2017explainable}, unsupervised learning~\cite{hinton1999unsupervised}, reinforcement learning~\cite{wiering2012reinforcement}, application to graphs resp. dynamically structured data~\cite{wang2019knowledge}, computing metrics~\cite{kulis2012metric}, autonomous learning or artificial intelligence~\cite{goertzel2007artificial}, and many more. In this work, we deal with an alternative to residual blocks that optimize the function $R(x,y)=f(x)+y$. $x$ and $y$ are here two consecutive outputs of convolution layers.
	
	Our approach builds on maxout nets~\cite{goodfellow2013maxout}, which were introduced as activation functions requiring twice the number of neurons. These maxout nets optimize the function $M(x,y)=max(x,y)$ where $x$ and $y$ are the outputs of two neuron or convolution layers which are parallel to each other (Receiving the same input). This concept can be directly integrated as a replacement for the residual function, thus not requiring any additional learnable parameter or neurons. This means we use the function $M(x,y)=max(x,y)$ whereby $x$ and $y$ are consecutive convolutions. In addition, we formulate the leaky max propagation (LMP) to combine the residual function with the maxout function. This combination optimizes the equation $LMP(x,y)=max(x,y)*\alpha + min(x,y)*\beta$ where $x$ and $y$ are again sequential convolutional layer outputs and $\alpha$ and $\beta$ are weighting factors which are fixed before training and satisfy the condition $\alpha+\beta=1$. 
	
	\setlength{\tabcolsep}{0pt}
	\subsection{Difference to Maxout}
	\begin{itemize}
		\setlength{\itemsep}{-3pt}
		\item We do not need twice the amount of neurons
		\item We use Maxout to combine successive convolutional layers
		\item We use it as a combination layer and activation function
		\item We extended the formulation (Leaky Max Propagation) to propagate a portion of the residual formulation forward and backward through the network as well.
	\end{itemize}
	\subsection{Contribution of this work}
	\begin{itemize}
		\setlength{\itemsep}{-3pt}
		\item We show that maximum propagation can be used effectively as a replacement for residual connections as well as an alternative to form ensambles.
		\item We extend the formulation of maximum propagation to allow a fraction to pass (leaky maximum propagation).
		\item Our proposed approaches generalize well without batch normalization and since batch normalization makes DNNs less robust~\cite{benz2021revisiting,benz2020batch}, we see this as an advantage.
		\item Our proposed approaches can be trained without additional activation function.
	\end{itemize}

	\section{Method}
	
	\begin{figure}[htb]
		\centering
		\includegraphics[width=0.95\textwidth]{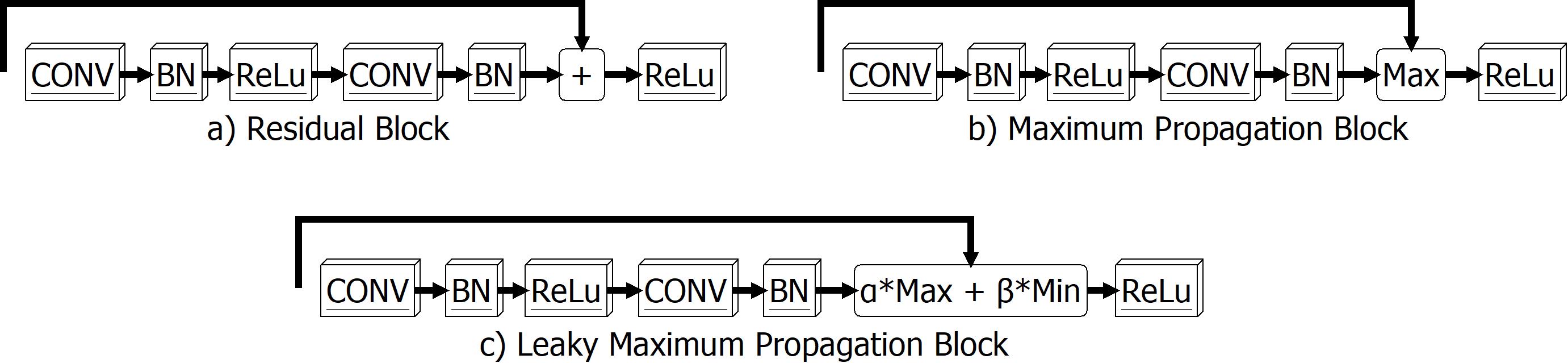}
		\caption{Visual description of the architecture from residual blocks (a), maximum propagation blocks (b), and leaky maximum propagation blocks (c) for ResNet-34 (ResNet-50 uses two depthwise convolution before the $3 \times 3$ convolution block in the center but the concept is identical). All three share the same placement of the convolution (CONV), batch normalization (BN), and rectifier linear unit (ReLu) blocks. The only difference is the performed operation to combine the result of the block computation and the input to the block.}
		\label{fig:blocks}
	\end{figure}
	
	Figure~\ref{fig:blocks} shows a visual representation of the structure of residual blocks (a), maximum propagation blocks (b), and leaky maximum propagation blocks (c). There is no difference between the blocks except for the final combination of the block's input and the output of the internal layers. Formally, the residual blocks and the maximum propagation blocks optimize different functions where the leaky maximum propagation function is a generalization of both functions. For the input of a block $x \in \R^{d}$ and the function of the internal blocks $f(x)$, the function which is optimized in residual blocks is given by:
	
	\begin{equation}
		Residual(x) = f(x) + x
		\label{eq:residual}
	\end{equation}
	
	The function~\ref{eq:residual} tries to find an optimization $f(x)$ for the input $x$. In the case of blocks connected in series, this means that each block produces an improvement for the previous output, which is passed from block to block. This results in a nested optimization that can be interpreted as an ensemble of many small networks~\cite{veit2016residual}. The biggest advantage of residual blocks, however, is that they help circumvent the vanishing gradient problem~\cite{ veit2016residual}.
	
	\begin{equation}
		MaxProp(x) = \left\{\begin{array}{ll} f(x), & f(x) \geq x \\
			x, & else \end{array}\right. 
		\label{eq:maxprop}
	\end{equation}
	
	In equation~\ref{eq:maxprop}, the maximum propagation function is shown, which is also used as the activation function in maxout nets~\cite{goodfellow2013maxout}. Since we use this as a combining function for nested blocks, we do not need twice the amount of neurons as is the case in maxout nets~\cite{goodfellow2013maxout}. Compared to the residual blocks, only the maximum value is forwarded. Hereby, the whole network optimizes different long and small nets in parallel. This means that whole blocks can be skipped internally. For the backpropagation of the error, both inputs receive the same error value, which is added to the input connection since it receives the error of the internal function additionally.

	\begin{figure}[htb]
		\centering
		\includegraphics[width=0.95\textwidth]{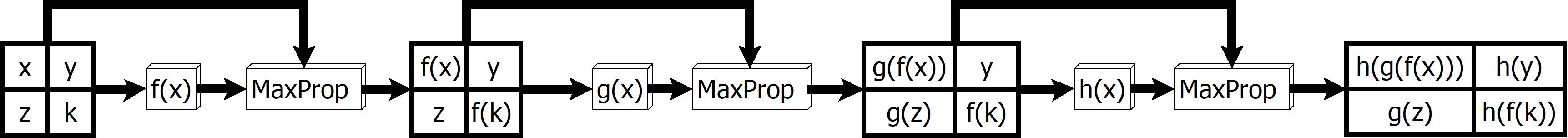}
		\caption{Visual description of functions learned by consecutive maximum propagation blocks. On the left side is the input to the three blocks with the internal functions f(x), g(x), and h(x). Based on the maximum selection each input has a different output function.}
		\label{fig:jumpintern}
	\end{figure}
	
	Figure~\ref{fig:jumpintern} shows a visual representation of the function approximation of maximum propagation blocks. Each input value (x,y,z,k) is assigned a different output function based on the maximum selection. The functions f(x), g(x), and h(x) are computed by the internal convolutions, batch normalization, and activation functions in the intermediate layer as shown in Figure~\ref{fig:blocks}. Through this, the network now no longer learns an optimization to the previous layer in each layer, but different deep networks, which are assembled based on the maximum values. For the backpropagation, the error is given to the maximum value only, which means, that the lesser value does not receive an error and therefore, can be theoretically unused in the network. This brings us to our general formulation of the maximum propagation, which is the leaky maximum propagation function in equation~\ref{eq:leakymaxprop}.
	
	\begin{equation}
		MaxProp(x) = \left\{\begin{array}{ll} f(x)*\alpha + x*\beta, & f(x) \geq x \\
			x*\alpha + f(x)*\beta, & else \end{array}\right. 
		\label{eq:leakymaxprop}
	\end{equation}
	
	The leaky maximum propagation function has two additional parameters $\alpha$ and $\beta$, which have to be set in advance. These two parameters, specify how much weight is given to the maximum and the minimum value. For the parameters $\alpha=1.0$ and $\beta=1.0$ this would be the residual function. If we use $\alpha=1.0$ and $\beta=0.0$ we would have the maximum propagation function. For any other value, we receive a function, which optimizes both functions in parallel but weighted (With the exception of $\alpha=0.0$ and $\beta=0.0$). This means that the leaky maximum propagation function is a generalization of the residual and maximum propagation function. For the back propagation of the error we use the same values for $\alpha$ and $\beta$ to assign each input the weighted error. The idea behind this formulation is to overcome a possible unused layer which can happen in the maximum propagation function.
	
	\begin{figure}[htb]
		\centering
		\includegraphics[width=0.95\textwidth]{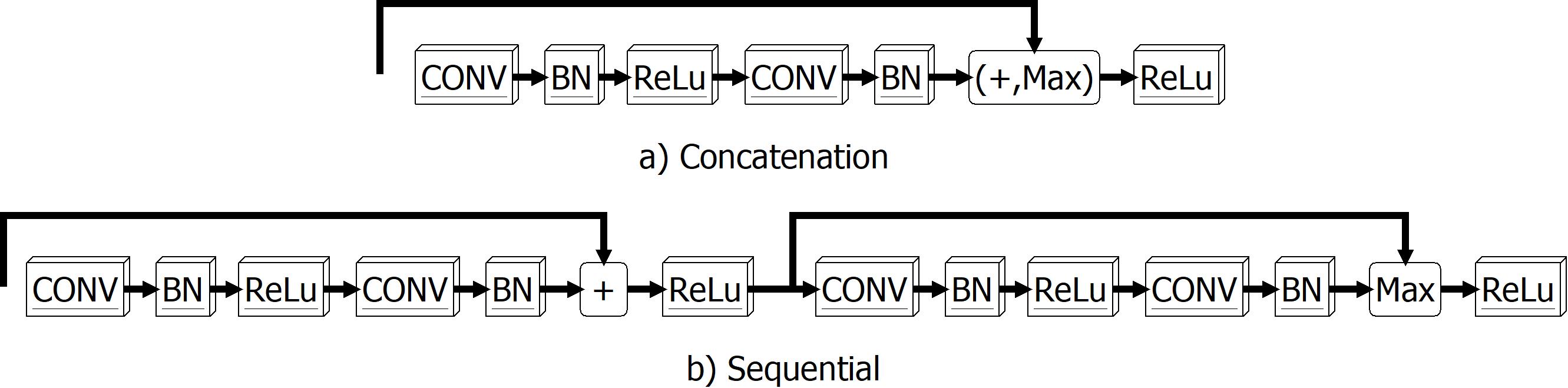}
		\caption{Visual description of possible alternatives to combine the residual and maximum propagation layer. In a) the concatenation is shown and in b) the alternating interconnection of the layers.}
		\label{fig:combies}
	\end{figure}
	
	Figure~\ref{fig:combies} shows further possible combinations of the residual and maximum propagation blocks as an alternative to the leaky maximum propagation formulation. In Figure~\ref{fig:combies} a) the results of the maximum propagation and the residual blocks are concatenated. Hereby, each layer has only half the depth (Or output channels) to allow a fair comparison with the same number of parameters. This is the use of the inception~\cite{szegedy2015going} technique which processes several small nets in parallel and concatenates their results at the end. Figure~\ref{fig:combies} b) shows the residual and maximum propagation blocks connected in series. Here the residual block optimizes the previous result and the maximum propagation block selects either the improvement or another internal optimization. These nestings can be continued arbitrarily. However, this requires a large number of GPUs and a grid search over many possibilities, which is beyond the scope of this work. In this work, we consider only the alternating use of residual and maximum propagation blocks if "Alternating Blocks" is specified in brackets after a model.
	
	\begin{figure}[htb]
		\centering
		\includegraphics[width=0.95\textwidth]{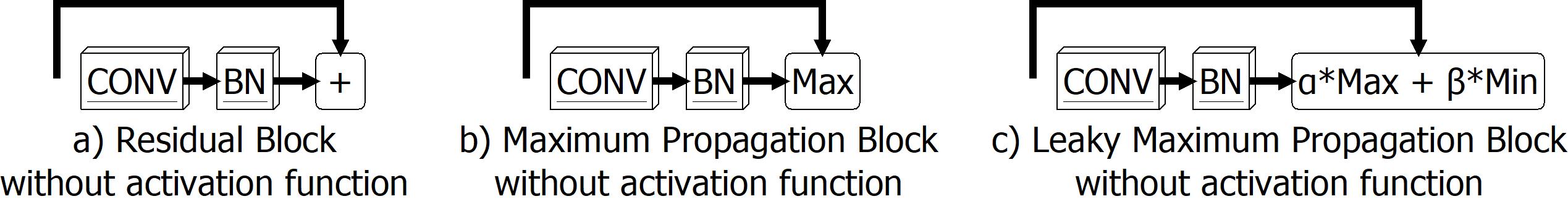}
		\caption{Visual description of the architecture from residual blocks (a), maximum propagation blocks (b), and leaky maximum propagation blocks (c) without any actiavation function for ResNet-34 blocks (ResNet-50 would requier to pack each convolution block between a max propagation connection). All three share the same placement of the convolution (CONV) and batch normalization (BN) blocks. The only difference is the performed operation to combine the result of the block computation and the input to the block. There is also no activation function after the block and due to the global average pooling~\cite{lin2013network} no activation function in the entire model.}
		\label{fig:blocksNOactivation}
	\end{figure}
	
	Figure~\ref{fig:blocksNOactivation} shows the use of the residual (a), maximum propagation (b), and leaky maximum propagation (c) blocks without activation function. For this, the two internal convolutions must also be reduced to one, doubling the number of layers, in order to arrive at the same set of learnable parameters and allow for a fair comparison. The ReLu after each block was also removed and since we used global average pooling~\cite{lin2013network} and only one fully connected layer at the end, there is no activation function in the whole network. This is a very interesting property of maximum propagation, since maximum selection is already an activation function, as shown earlier in maxout nets~\cite{goodfellow2013maxout}. For residual blocks (Figure~\ref{fig:blocksNOactivation} a)), this does ofcourse not work and the network will either learn nothing or end up in not a number. For maximum propagation this work surprisingly and even better than the leaky maximum propagation, which we will show in our results.

	\section{Data sets}
	In this section we describe all used data sets and the training and test splits we used in this work.
	
	\textbf{SVHN}~\cite{netzer2011reading} is a public data set with real world images from house numbers. The data set contains the tasks number detection and number classification. In our evaluation we only used the classification of the provided cantered single characters. Similar to MNIST~\cite{lecun1998mnist} it contains all numbers from 0 to 9. The resolution of each image is $32 \times 32$ with all color channels (red, green, and blue). For training the authors provide 73,257 images and for validation 26,032 images. There is also a third set without annotations which we did not use in our evaluation.
	
	\textbf{FashionMNIST}~\cite{xiao2017online} is an image classification data set inspired by MNIST~\cite{lecun1998mnist}. The images are gray scale and have a resolution of $28 \times 28$. Instead of the original letters from the MNIST data set the authors use the classes T-shirt, Trouser, Pullover, Dress, Coat, Sandal, Shirt, Sneaker, Bag, and Boot. The data set contains 60,000 images for training and 10,000 images for validation. 
	
	\textbf{CIFAR10}~\cite{krizhevsky2009learning} is a publicly available data set with RGB (red, green, and blue) images and 10 classes. The resolution of the images is $32 \times 32$ and the authors provided 50,000 images for training and 10,000 images for validation. The data set is balanced which means, that for each class the same amount of samples is provided.
	
	\textbf{CIFAR100}~\cite{krizhevsky2009learning} is a similar data set to CIFAR10 but with 100 instead of 10 classes. The resolution of the images is $32 \times 32$ and each image has all three color channels (red, green, and blue). The training set contains 500 images per class which sums up to 50,000 images. The validation set has 100 images per class which is 10,000 in total. As CIFAR10 this data set is balanced.
	
	\textbf{ImageNet ILSVRC2015}~\cite{krizhevsky2012imagenet} is one of the largest publicly available image classification data sets. It has one thousand classes and more than one million images for training. It can be used for object detection and image classification. In our evaluation we only used the classification task. The images have different resolutions which contains an additional challenge. For validation 50,000 images are provided.
	
	\textbf{VOC2012}~\cite{pascalvoc2012} is a publicly available data set for semantic segmentation. In addition to the semantic segmentation it also contains annotations for object detection and object classification. In our evaluation we only used the semantic segmentation. Semantic segmentation means that the object has to be extracted from the image with pixel accuracy and each pixel has to be classified into the object class. Each image in VOC2012 can contain multiple objects and the task is to segment them as well as classify each pixel correctly. For training 1,464 images are provided with 3,507 segmented and classified objects. The validation set consists of 1,449 images with 3,422 segmented objects. In addition to the images with segmented objects, a third set is provided without annotations. In our evaluation we did not use the third set.
	
	\textbf{500kCloser}~\cite{CAIP2019FuhlW,NNETRA2020} is a publicly available data set with 866,069 images. It contains semantic segmentations for the pupil and the iris as well as eyeball parameters and gaze vectors. In our evaluation we only used the regression of the eyeball parameters as well as the gaze vector. The resolution of the images is $192 \times 144$ and the recordings are gray scale due to the near infrared illumination used. The images are split in two videos, one from simulator driving recordings (509,379 images) and one from real world driving recordings (356,690 images). Since there is no predefined train and test split we used the simulator recordings for training and the real-world recordings for validation.

	\section{Evaluation}
	\label{sec:eval}
	In this section we show the performance of maximum propagation and leaky maximum propagation in comparison to residual blocks, which are marked as (Addition) in the evaluations, on a variety of publicly available data sets. For the leaky maximum propagation, we always used $\alpha=0.9$ and $\beta=0.1$. The hardware to train all the models is an AMD Ryzen 9 2950X with 3.50 GHz and 64 GB DDR4 Ram. The used GPU is an NVIDIA 3090 RTX with 24 GB DDR6X Ram. As deep learning framework we used DLIB~\cite{king2009dlib} version 19.21 in which the provided code can be copy pasted an directly executed. The NVIDIA driver version of the system is 461.09, the operating system is Windows 10 64Bit, the CUDA version is 11.2, and the cuDNN version used is 8.1.0.
	
	\begin{table}[htb]
		\caption{Shows the classification accuracy results for the data sets CIFAR10, CIFAR100, FashionMNIST, and SVHN with different architectures and \textbf{batch normalization} as average over 5 runs with standard deviation after $\pm$. The text in the brackets after the architecture specifies the performed combination operation to the input and output of each block.\\
			\textit{Training parameters: Optimizer=SGD, Momentum=0.9, Weight Decay=0.0005, Learning rate=0.1, Batch size=100, Training time=100 epochs, Learning rate reduction after each 20 epochs by 0.1}\\
			\textit{Data augmentation: Shifting by up to 4 pixels in each direction and padding with zeros. Mean (Red=122, Green=117, Blue=104) subtraction and division by 256.}}
		\label{tbl:CLassSmallBN}
		\centering
		\begin{tabular}{lcccc}
			\toprule
			Method & CIFAR10 & CIFAR100 & FashionMNIST & SVHN \\
			\midrule
			ResNet-34 (Additon) & $92.52 \pm 0.25$ & $73.16 \pm 0.61$ & $94.83 \pm 0.22$ & $96.1 \pm 0.23$ \\
			ResNet-34 (Maximum) & $93.36 \pm 0.08$ & $73.13 \pm 0.69$ & $95.00 \pm 0.18$ & $\bm{96.36 \pm 0.09}$ \\
			ResNet-34 (Leaky Max) & $\bm{93.88 \pm 0.09}$ & $\bm{73.66 \pm 0.2}$ & $\bm{95.05 \pm 0.14}$ & $96.19 \pm 0.19$ \\ \hline
			ResNet-50 (Additon) & $93.13 \pm 0.19$ & $74.41 \pm 0.41$ & $95.44 \pm 0.19$ & $96.77 \pm 0.19$ \\
			ResNet-50 (Maximum) & $94.02 \pm 0.21$ & $74.4 \pm 0.41$ & $95.77 \pm 0.21$ & $\bm{97.11 \pm 0.21}$ \\
			ResNet-50 (Leaky Max) & $\bm{94.49 \pm 0.18}$& $\bm{74.91 \pm 0.4}$ & $\bm{96.01 \pm 0.18}$ & $96.95 \pm 0.21$ \\
			\bottomrule
		\end{tabular}
	\end{table}
	
	Table~\ref{tbl:CLassSmallBN} shows the results with batch normalization for the model ResNet-34 and ResNet-50. Averaged over five runs, the maximum propagation and the leaky maximum propagation seem to perform at least as well or even slightly better than ordinary residual connections. Between the leaky maximum propagation and the maximum propagation only the data set SVHN seems to make a difference, whereas the maximum propagation seems to work a bit better. Since we can't see much difference, although the maximum propagation networks learn a different function (See Figure~\ref{fig:jumpintern}), with the standard networks, we investigated why this is the case and found that batch normalization does a significant amount of the work for residual connections. 
	
	\begin{table}[htb]
		\caption{Shows the classification accuracy results for the data sets CIFAR10, CIFAR100, FashionMNIST, and SVHN with different architectures and \textbf{fixed batch normalization (Mean=0,Std=1,Scale=1,Shift=0)} as average over 5 runs with standard deviation after $\pm$. The text in the brackets after the architecture specifies the performed combination operation to the input and output of each block.\\
			\textit{Training parameters: Optimizer=SGD, Momentum=0.9, Weight Decay=0.0005, Learning rate=highest possible (0.01 or 0.001), Batch size=100, Training time=100 epochs, Learning rate reduction after each 20 epochs by 0.1}\\
			\textit{Data augmentation: Shifting by up to 4 pixels in each direction and padding with zeros. Mean (Red=122, Green=117, Blue=104) subtraction and division by 256.}}
		\label{tbl:CLassSmallAFF}
		\centering
		\begin{tabular}{lcccc}
			\toprule
			Method & CIFAR10 & CIFAR100 & FashionMNIST & SVHN \\
			\midrule
			ResNet-34 (Additon) & $76.83 \pm 0.72$ & $35.63 \pm 0.2$ & $92.4 \pm 0.15$ & $93.16 \pm 0.05$ \\
			ResNet-34 (Maximum) & $\bm{88.76 \pm 0.05}$ & $\bm{59.96 \pm 0.23}$ & $\bm{93.45 \pm 0.09}$ & $\bm{95.92 \pm 0.05}$ \\
			ResNet-34 (Leaky Max) & $80.26 \pm 0.1$ & $44.54 \pm 0.46$ & $93.09 \pm 0.15$ & $94.45 \pm 0.02$ \\
			\bottomrule
		\end{tabular}
	\end{table}
	
	Table~\ref{tbl:CLassSmallAFF} shows the results with a constant batch normalization so the parameters Mean=0, Std=1, Scale=1, and Shift=0 are constant during the whole training process and also during the validation. Here it becomes clear that the maximum propagation generalizes best, which is especially the case for CIFAR100 and CIFAR10. Another interesting property of the constant batch normalization, is the reduction of the variance of the results for the maximum propagation. This clearly shows that the maximum propagation as well as the leaky maximum propagation learn different functions compared to residual connections. Of course, this also raises the question of why not use batch normalization. Here we would mention firstly the reduced robustness of networks caused by batch normalization~\cite{benz2021revisiting,benz2020batch}. Since our approaches do not require batch normalization for generalization, we see this as an advantage of our approaches. Another point in favor of removing batch normalization is that it reduces the complexity of the networks and thus simplifies validation in future research.
	
	\begin{table}[htb]
		\caption{Shows the classification accuracy results for the data sets CIFAR10, CIFAR100, FashionMNIST, and SVHN with different architectures and \textbf{batch normalization} as average over 5 runs with standard deviation after $\pm$. The text in the brackets after the architecture specifies the performed combination operation to the input and output of each block. For this evaluation we focused on different combination approaches like concatenation and the useage of alternating blocks in the same architecture. Alternating in this context means, one maximum propagation block is always followed by one residual block.\\
			\textit{Training parameters: Optimizer=SGD, Momentum=0.9, Weight Decay=0.0005, Learning rate=0.1, Batch size=100, Training time=100 epochs, Learning rate reduction after each 20 epochs by 0.1}\\
			\textit{Data augmentation: Shifting by up to 4 pixels in each direction and padding with zeros. Mean (Red=122, Green=117, Blue=104) subtraction and division by 256.}}
		\label{tbl:CLassSmallBNApproach}
		\centering
		\begin{tabular}{lcccc}
			\toprule
			Method & CIFAR10 & CIFAR100 & FashionMNIST & SVHN \\
			\midrule
			ResNet-34 (Leaky Max) & \textbf{$\bm{93.88 \pm 0.09}$} & \textbf{$\bm{73.66 \pm 0.2}$} & \textbf{$\bm{95.05 \pm 0.14}$} & \textbf{$\bm{96.19 \pm 0.19}$} \\
			ResNet-34 (Concatenation) & $92.56 \pm 0.24$ & $71.38 \pm 0.54$ & $91.52 \pm 0.64$ & $95.02 \pm 0.37$ \\
			ResNet-34 (Alternating) & $93.04 \pm 0.35$ & $72.39 \pm 0.4$ & $95.06 \pm 0.36$ & $96.06 \pm 0.1$ \\ \hline \hline
			ResNet-50 (Leaky Max) & \textbf{$\bm{94.49 \pm 0.18}$} & \textbf{$\bm{74.91 \pm 0.4}$} & \textbf{$\bm{96.01 \pm 0.18}$} & \textbf{$\bm{96.95 \pm 0.21}$} \\
			ResNet-50 (Concatenation) & $93.18 \pm 0.2$ & $72.65 \pm 0.4$ & $92.21 \pm 0.21$ & $95.63 \pm 0.18$ \\
			ResNet-50 (Alternating) & $93.68 \pm 0.21$ & $73.67 \pm 0.4$ & $95.68 \pm 0.21$ & $96.75 \pm 0.21$ \\
			\bottomrule
		\end{tabular}
	\end{table}
	
	Table~\ref{tbl:CLassSmallBNApproach} shows the results for different possible combinations of residual connections and maximum propagation. By "concatenation" we mean the concatenation of the addition and the maximum as shown in Figure~\ref{fig:combies} a). For these nets we used the inception model where each sub-block had only half the depth to get the same number of parameters as all other nets. By "alternating" we refer to the alternating use of residual blocks and maximum propagation as shown in Figure~\ref{fig:combies} b). Based on the results in Table~\ref{tbl:CLassSmallBNApproach}, it is evident that the leaky formulation is the most effective over all data sets.
	
	\begin{table}[htb]
		\caption{Shows the classification accuracy results for the data sets CIFAR10, CIFAR100, FashionMNIST, and SVHN with ensambles of six ResNet-34 architectures. The text in the brackets after the architecture specifies the amount of models per approch is in the ensamble (A=Residual blocks or Additon, M=Maximum propagation, LM=Leaky maximum propagation).\\
			\textit{Training parameters: Optimizer=SGD, Momentum=0.9, Weight Decay=0.0005, Learning rate=0.1, Batch size=100, Training time=100 epochs, Learning rate reduction after each 20 epochs by 0.1}\\
			\textit{Data augmentation: Shifting by up to 4 pixels in each direction and padding with zeros. Mean (Red=122, Green=117, Blue=104) subtraction and division by 256.}}
		\label{tbl:Ensamble}
		\centering
		\begin{tabular}{lcccc}
			\toprule
			Method & CIFAR10 & CIFAR100 & FashionMNIST & SVHN \\
			\midrule
			ResNet-34 (6*A,0*M,0*LM) & 93.64\% & 75.42\% & 95.18\% & 96.14\% \\
			ResNet-34 (0*A,6*M,0*LM) & 94.97\% & 75.59\% & 95.46\% & 96.89\% \\
			ResNet-34 (0*A,0*M,6*LM) & 94.44\% & 74.94\% & 95.19\% & 96.67\% \\ \hline
			ResNet-34 (3*A,3*M,0*LM) & 95.36\% & 78.02\% & 95.61\% & 96.68\% \\
			ResNet-34 (3*A,0*M,3*LM) & 94.65\% & 77.15\% & 95.49\% & 96.49\% \\
			ResNet-34 (0*A,3*M,3*LM) & 95.39\% & 77.83\% & 95.81\% & 97.13\% \\ \hline
			ResNet-34 (2*A,2*M,2*LM) & \textbf{95.99\%} & \textbf{78.55\%} & \textbf{95.99\%} & \textbf{97.96\%} \\
			\bottomrule
		\end{tabular}
	\end{table}
	
	Table~\ref{tbl:Ensamble} shows the results for the formation of ensembles. Here, of course, it is important that networks learn different functions to benefit from the combination. This can be seen in the central section of Table~\ref{tbl:Ensamble} in which almost all combinations are significantly better than combining the same nets (First section in Table~\ref{tbl:Ensamble}). Of course, there is an exception here, which is the maximum propagation for the data set SVHN and CIFAR10. In these two cases, combining the same nets with the maximum propagation is slightly better in comaprison to some other combinations. In the last section in Table~\ref{tbl:Ensamble}, all approaches are combined and also give the best results overall.
	
	\begin{figure}[htb]
		\centering
		\includegraphics[width=0.48\textwidth]{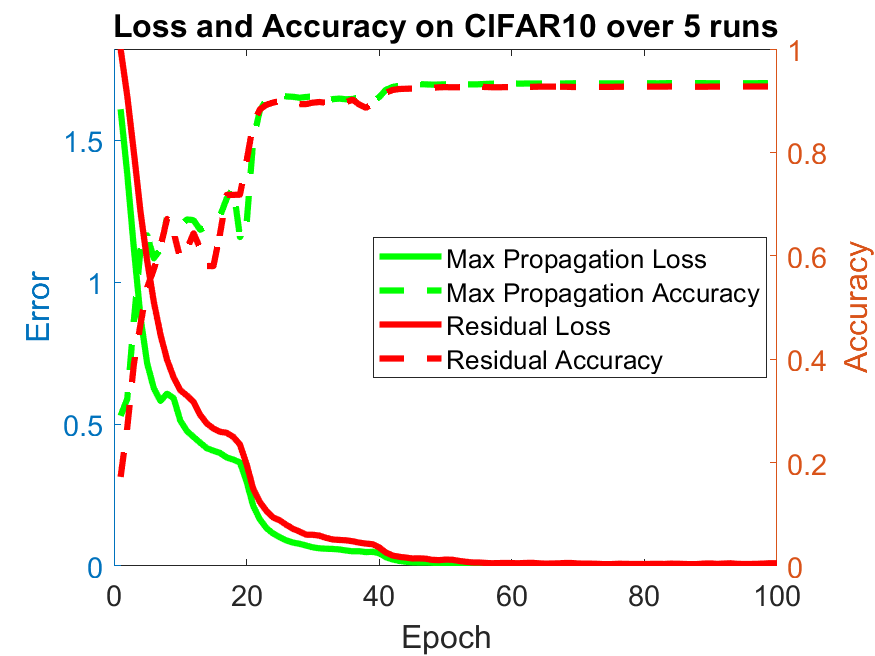}
		\includegraphics[width=0.48\textwidth]{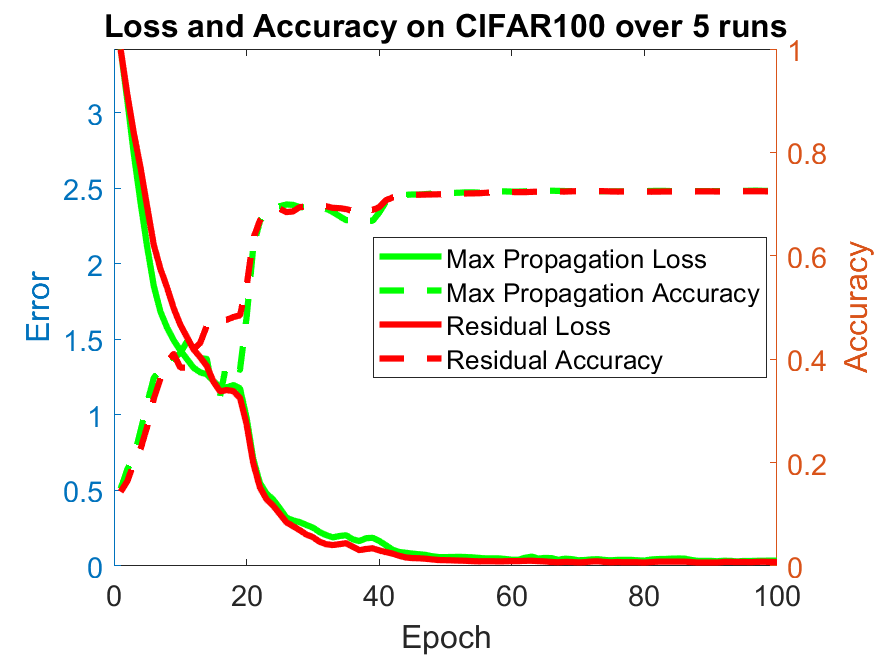}
		\includegraphics[width=0.48\textwidth]{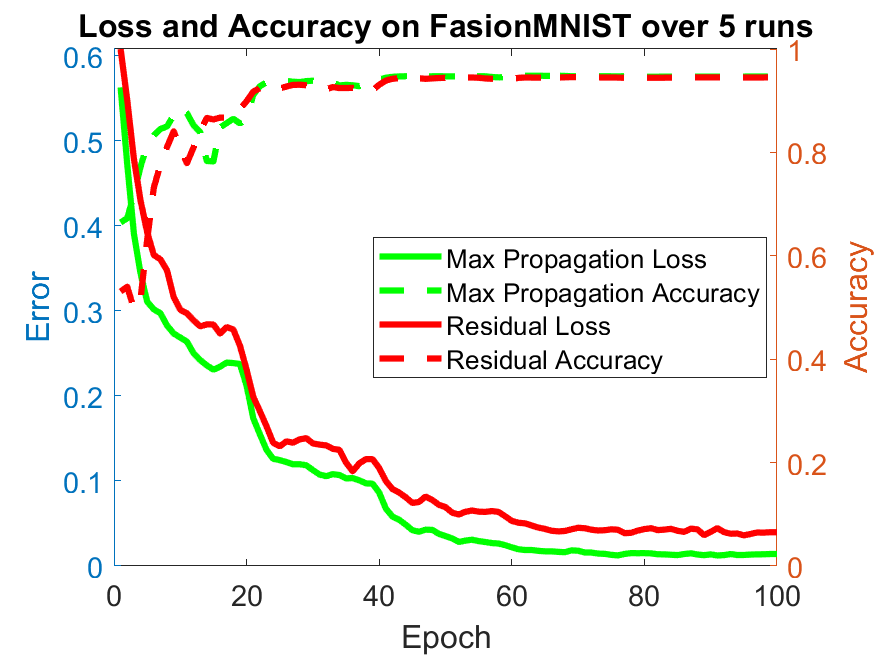}
		\includegraphics[width=0.48\textwidth]{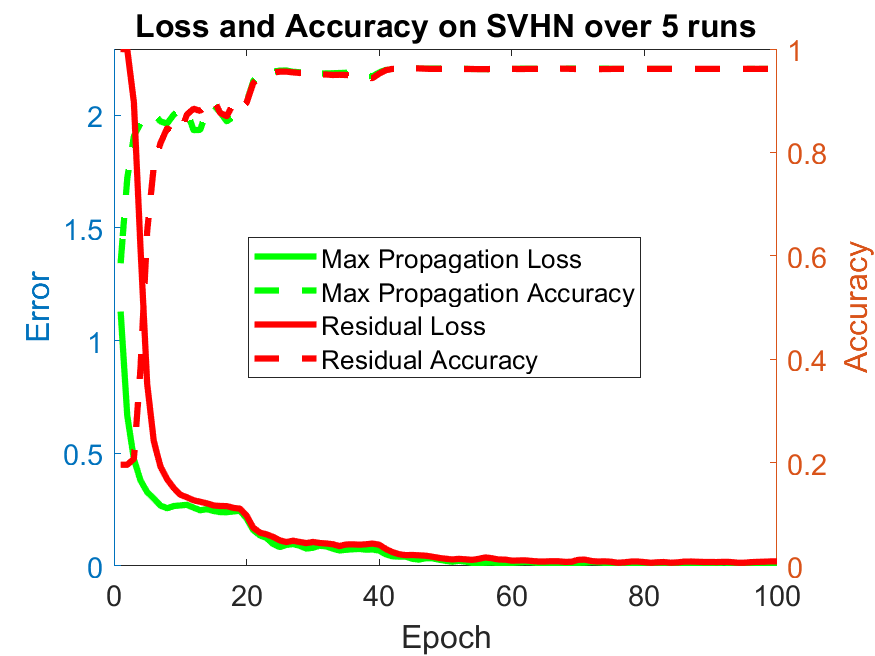}
		\caption{Visual progress of the loss and accuracy as average over 5 runs for CIFAR10, CIFAR100, FashionMNIST, and SVHN. The solid lines are the loss values and the dashed lines are the accuracy. Red represents the values for the residual blocks and green the maximum propagation. The used architecture is ResNet-34 and every 20 epochs the learning rate was reduced by 0.1.}
		\label{fig:loss}
	\end{figure}
	
	Figure~\ref{fig:loss} shows the loss (Solid line) and the accuracy (Dashed line) for the maximum propagation (Green) and residual connections (Red) over five runs as average value. As can be seen, the maximum propagation reduces the loss faster and therefore requires less time to learn. Note that after each 20 epochs the learning rate was reduced by 0.1. An exception here is the second segment (Epoch 20 to 40) for the CIFAR100 data set. This shows that the faster learning is not always the case and as described earlier, the maximum propagation and leaky maximum propagation can be seen as an alternative to residual connections but not as a replacement which works always better.
	
	\begin{table}[htb]
		\caption{Shows the classification accuracy results for the data sets CIFAR10, CIFAR100, FashionMNIST, and SVHN with \textbf{batch normalization and no activation function} as average over 5 runs with standard deviation after $\pm$. The text in the brackets after the architecture specifies the performed combination operation to the input and output of each block. The used blocks can be seen in Figure~\ref{fig:blocksNOactivation}.\\
			\textit{Training parameters: Optimizer=SGD, Momentum=0.9, Weight Decay=0.0005, Learning rate=0.1, Batch size=100, Training time=100 epochs, Learning rate reduction after each 20 epochs by 0.1}\\
			\textit{Data augmentation: Shifting by up to 4 pixels in each direction and padding with zeros. Mean (Red=122, Green=117, Blue=104) subtraction and division by 256.}}
		\label{tbl:CLassSmallBNNoActivation}
		\centering
		\begin{tabular}{lcccc}
			\toprule
			Method & CIFAR10 & CIFAR100 & FashionMNIST & SVHN \\
			\midrule
			ResNet-34 (Additon) & nan & nan & nan & nan \\
			ResNet-34 (Maximum) & $\bm{93.78 \pm 0.13}$ & $\bm{71.08 \pm 2.29}$ & $94.82 \pm 0.27$  & $ 96.19 \pm 0.11$ \\ 
			ResNet-34 (Leaky Maximum) & $87.69 \pm 2.1$ & $74.04 \pm 0.25$ & $\bm{95.15 \pm 0.17}$ & $96.39 \pm 0.16$  \\
			\bottomrule
		\end{tabular}
	\end{table}

	Table~\ref{tbl:CLassSmallBNNoActivation} shows the results without additional activation function in or after a residual, maximum, or leaky maximum connection. This is also visualized in Figure~\ref{fig:blocksNOactivation}. For the residual connection this does not work since it ends up in not a number indicated by nan. For the maximum and leaky maximum propagation it still works but based on the random initialization of the network it can have a negative impact on the results as can be seen for CIFAR10 for the leaky maximum propagation as well as for CIFAR100 for the maximum propagation. In the following, we show that the presented maximum propagation and leaky maximum propagation also work for large data sets like ImageNet, and that they can also be used for semantic segmentation and regression.
	
	\begin{table}[htb]
		\caption{Shows the classification accuracy results for the data set ILSVRC2015 or ImageNet. The text in the brackets after the architecture specifies the performed combination operation to the input and output of each block.\\
			\textit{Training parameters: Optimizer=SGD, Momentum=0.9, Weight Decay=0.0001, Learning rate=0.1, Batch size=160, Training time=1500 epochs, Learning rate reduction after each 500 epochs by 0.1}\\
			\textit{Data augmentation: Random cropping of $227 \times 227$ regions. Random flipping in each direction. Random color offset. Mean (Red=122, Green=117, Blue=104) subtraction and division by 256.}}
		\label{tbl:CLassImageNet}
		\centering
		\begin{tabular}{lcc}
			\toprule
			Method & Top-1 & Top-5 \\
			\midrule
			ResNet-34 (Additon) & 75.33\% & 92.43\% \\
			ResNet-34 (Maximum) & 75.23\% & 92.31\% \\
			ResNet-34 (Leaky Maximum) & \textbf{75.55\%} & \textbf{92.90\%} \\
			\bottomrule
		\end{tabular}
	\end{table}
	
	Table~\ref{tbl:CLassImageNet} shows the Top-1 and Top-5 accuracy on ImageNet. The maximum propagation has the lowest results. The best results are obtained by the leaky maximum propagation. Since the training and evaluation on ImageNet consume a lot of time, we made only one training and evaluation run. Therefore, we think, that the results show that all three approaches are comparable since for each training and evaluation run the results change slightly as it is the case for all of the other data sets.

	\begin{table}[htb]
		\caption{Shows the pixel classification accuracy of the semantic segmentation results for the data set VOC2012. The text in the brackets after the architecture specifies the performed combination operation to the input and output of each block.\\
			\textit{Training parameters: Optimizer=SGD, Momentum=0.9, Weight Decay=0.0005, Learning rate=0.1, Batch size=30, Training time=800 epochs, Learning rate reduction after each 400 epochs by 0.1}\\
			\textit{Data augmentation: Random cropping of $227 \times 227$ regions. Random flipping in each direction. Random color offset. Mean (Red=122, Green=117, Blue=104) subtraction and division by 256.}}
		\label{tbl:Segi}
		\centering
		\begin{tabular}{lc}
			\toprule
			Method & Pixel Accuracy  \\
			\midrule
			U-Net (Additon) & 85.15\%  \\
			U-Net (Maximum) & 85.34\%\\ 
			U-Net (Leaky Maximum) & \textbf{85.58\%}\\
			\bottomrule
		\end{tabular}
	\end{table}
	
	Table~\ref{tbl:Segi} shows the pixel accuracy on VOC2012 for the semantic segmentation task. As can be seen the maximum propagation and leaky maximum propagation work slightly better in comparison to the residual connection but since it is evaluated in a single run this cannot be seen as an evident result. We just want to show, that the proposed approaches work for semantic segmentation too.

	\begin{table}[htb]
		\caption{Shows the mean absolute distance for the eyeball center, eyeball radius, and optical vector on the results for the data set 500kCloser. The text in the brackets after the architecture specifies the performed combination operation to the input and output of each block.\\
			\textit{Training parameters: Optimizer=SGD, Momentum=0.9, Weight Decay=0.0005, Learning rate=0.001, Batch size=10, Training time=1000 epochs, Learning rate reduction after each 300 epochs by 0.1}\\
			\textit{Data augmentation: Random noise up to 20\%. Random image shift of up to 20\% of the image. Random image blurring with $\sigma$ up to 1.5. Random image overlay with up to 30\% of its intensity. Random color offset. Random color offset. Mean (Gray-scale=125) subtraction and division by 256.}}
		\label{tbl:Regression}
		\centering
		\begin{tabular}{lccc}
			\toprule
			Method & Center & Eyeball Radius & Optical Vector \\
			\midrule
			ResNet-34 (Additon) & $0.52\pm2.20$ & $\bm{0.03\pm0.02}$ & $0.20\pm0.10$  \\
			ResNet-34 (Maximum) & $0.66\pm2.37$ & $\bm{0.03\pm0.02}$ & $\bm{0.19\pm0.11}$  \\
			ResNet-34 (Leaky Maximum) & $\bm{0.52\pm2.10}$ & $0.04\pm0.02$ & $0.24\pm0.12$  \\
			\bottomrule
		\end{tabular}
	\end{table}
	
	Table~\ref{tbl:Regression} shows the regression results on the 500kCloser data set. As can be seen the residual connections as well as the maximum and leaky maximum propagation deliver more or less the same results. Since the data set is huge we also computed only a single run. Overall, Table~\ref{tbl:Regression} shows, that the proposed approaches work for regression with competitive results to residual connections.

	\begin{table}[htb]
		\caption{Shows the classification accuracy results for the data set CIFAR100 with different large DNN models. The \textbf{jointly trained ensamble (JTE)} consists of three subnetworks with addition, maximum propagation, and leaky maximum propagation connections, which are concatenated. A detailed description is given in the supplementary material. The number after the jointly trained ensamble stands for the amount of JTEs we have combined in a major voting ensamble. The difference between JTE1 and JTE2 is only the depth of the convolutions layers which of course also effects the size of the concatenated tensor before the last fully connected layer. \\
			\textit{Training parameters: Optimizer=SGD, Momentum=0.9, Weight Decay=0.0005, Learning rate=0.1, Batch size=50, Training time=150 epochs, Learning rate reduction after each 30 epochs by 0.1}\\
			\textit{Data augmentation: Shifting by up to 4 pixels in each direction and padding with zeros. Horizontal flipping and mean (Red=122, Green=117, Blue=104) subtraction as well as division by 256.}}
		\label{tbl:LargeNets}
		\centering
		\begin{tabular}{lcc}
			Method & CIFAR100 & Params \\ \hline
			ResNet-152 (Additon) & \textbf{76.17\%} & 60M \\
			ResNet-152 (Maximum) & 72.80\% & 60M \\
			ResNet-152 (Leaky Max) & 75.12\% & 60M \\ \hline
			WideResNet-28-10 (Additon) & \textbf{78.04\%} & 36.5M \\
			WideResNet-28-10 (Maximum) & 74.57\% &  36.5M \\
			WideResNet-28-10 (Leaky Max) & 76.89\% &  36.5M \\ \hline
			JTE1 $\times 1$ & 70.45\% & 1M \\
			JTE1 $\times 3$ & 74.14\% & 3M \\
			JTE1 $\times 6$ & 75.67\% & 6M \\
			JTE1 $\times 9$ & 77.18\% & 9M \\
			JTE2 $\times 1$ & 73.41\% & 3.5M \\
			JTE2 $\times 3$ & 75.73\% & 10.5M \\
			JTE2 $\times 6$ & 78.15\% & 21M \\
			JTE2 $\times 9$ & \textbf{79.89\%} & 31.5M \\
		\end{tabular}
	\end{table}

	Table~\ref{tbl:LargeNets} shows the results of the standard addition connections compared to the maximum and leaky maximum propagation. As can be clearly seen, the presented connections drastically reduce the accuracy of large models. This is especially true for the maximum propagation. In comparison, the ensambles of jointly trained ensambles achieve the same or even better results as the large networks with less parameters. This confirms the results in Table~\ref{tbl:Ensamble} where the nets were trained independently and the individual nets were also significantly larger. A detailed architecture of the jointly trained ensambles can be found in our supplementary material.

	\section{Limitations}
	\label{sec:limit}
	A limitation of the use of the maximum propagation or the leaky maximum propagation is that the selection of the maximum requires an additional comparison in the implementation. Through this the calculation should be minimally slower than residual blocks. In the case of the leaky maximum propagation the multiplications with $\alpha$ and $\beta$ are added. If however no activation functions are used, which permits the use of the maximum propagation, then the maximum propagation as well as the leaky maximum propagation should be minimally faster or equally fast to compute. In addition, the maximum and leaky maximum propagation, does not work as well as residual blocks for large networks but it outperforms the sole use of residual blocks in jointly trained ensambles with a similar sturucture as vision transformers. 
	
	\section{Conclusion}
	In this work, we present an alternative to conventional residual connections which propagates the maximum value or a combination of the maximum and minimum value. We showed in several evaluations, that it is competitive to usually used residual blocks and can be effectively combined in neuronal network ensembles together with residual networks. In addition to the competitive results, the maximum propagation and leaky maximum propagation has some interesting properties. One is the better generalization with a fixed batch normalization (Or without) which should help in creating more robust networks since it has been shown, that batch normalization is vulnerable to adversarial attacks. Our proposed approaches can also be trained without additional activation function which could help in validation approaches since the network complexity reduces. This is especially true if the batch normalization is also removed. Another interesting property is that for most of the data sets, the proposed approaches learn faster in comparison to residual connections. Future research should investigate the robustness of maximum propagation and leaky maximum propagation blocks as well as the possibility of validating such networks. A further alternative is the usage of such blocks in architecture search.

	\bibliographystyle{plain}
	\bibliography{template}

\end{document}